\documentclass[11pt]{article}

\usepackage[final]{acl}

\usepackage{times}
\usepackage{latexsym}

\usepackage{pifont}
\usepackage{amsmath,amssymb}
\usepackage{booktabs}
\usepackage{multirow}

\usepackage[T1]{fontenc}

\usepackage[utf8]{inputenc}

\usepackage{microtype}

\usepackage{inconsolata}

\usepackage{graphicx}

\title{TAPO: Tool-Aware Policy Optimization via Credit Transfer for Multimodal Search Agents}

\author{
  Chengqi Dong$^{1,2}$, Chuhuai Yue$^{2}$, Hang He$^{2}$, Yandong Liu$^{1}$, Fenghe Tang$^{1}$,\\ 
  \textbf{S Kevin Zhou}$^{1}$, \textbf{Xiaohan Wang}$^{2}$, \textbf{Jiajun Chai}$^{2}$, \textbf{Guojun Yin}$^{2}$ \\[0.3em]
  {\normalfont\small $^1$University of Science and Technology of China \quad $^2$Meituan}
}

\begin{document}
\maketitle
\begin{abstract}

We identify and formally characterize credit misassignment as a systematic failure mode of GRPO in tool-augmented multimodal search agents: its uniform broadcast of trajectory-level advantages to all tokens causes valuable tool-use steps in failing trajectories to be penalized no differently from valueless ones.
We further empirically quantify the scale of this phenomenon. Over half of failing trajectories and failing tool-use actions exhibit correctable credit misassignment, demonstrating that the wasted training signal is both substantial and structurally exploitable.
Building on this insight, we propose Tool-Aware Policy Optimization (TAPO), which exploits the parameter-determinism property of information-acquisition tools: similar call parameters define equivalent information-acquisition actions and should therefore share comparable action credit. TAPO constructs counterfactual witnesses within the current training batch and compensates misassigned negative credit via confidence-gated conservative advantage correction. It requires no additional annotation, models, or sampling, and introduces negligible computational overhead.
Across multiple multimodal search benchmarks, TAPO delivers consistent, plug-and-play improvements over strong baselines for three mainstream RL algorithms (GRPO, GSPO, and SAPO).
Our code and models will be publicly released upon acceptance.
\end{abstract}

\section{Introduction}

\begin{figure}[t]
  \centering
  \includegraphics[width=\linewidth]{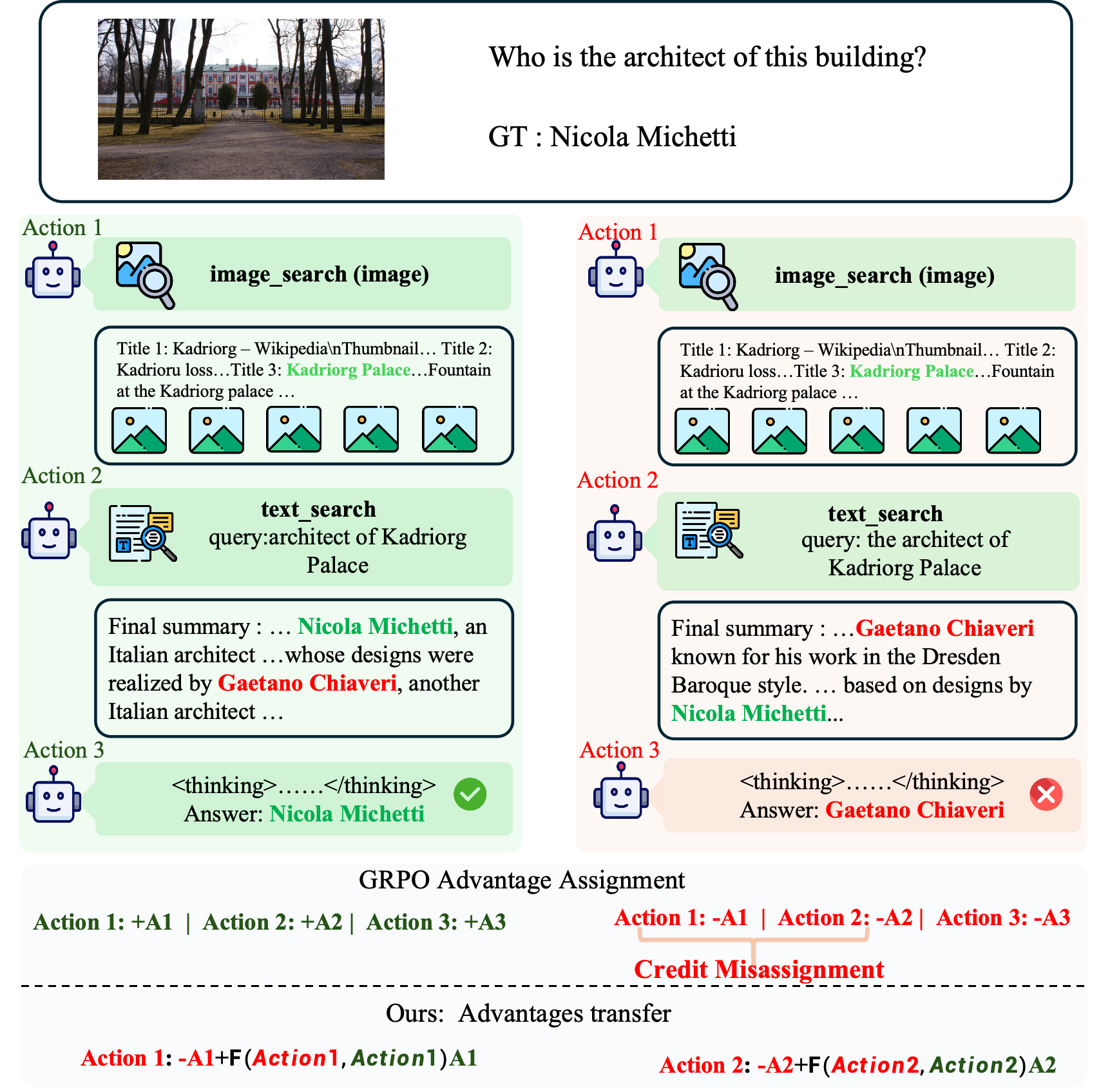}
  \caption{A concrete example of credit misassignment in group-based RL algorithms.
  }
  \label{fig:example}
\end{figure}

Agent-based reinforcement learning has emerged as the dominant paradigm for training tool-augmented multimodal agents,
enabling models to iteratively invoke external tools---such as text search, image search, and code interpreters---and
integrate tool outputs into coherent multi-step reasoning chains~\cite{zheng2025deepeyes,wu2025mmsearch, hong2025deepeyesv2, chng2025sensenova, dong2025training}.
Within this framework, Group Relative Policy Optimization (GRPO)~\cite{shao2024deepseekmath} has become the de facto training algorithm,
owing to its elimination of a separate value network and its strong empirical performance on outcome-supervised tasks.


GRPO uniformly broadcasts the trajectory-level advantage to every token, implicitly assuming equal contribution from each generation step---an assumption that systematically breaks down in multi-step tool-use scenarios where different tool use contribute different informational value.

\begin{figure*}[t]
  \centering
  \includegraphics[width=\textwidth]{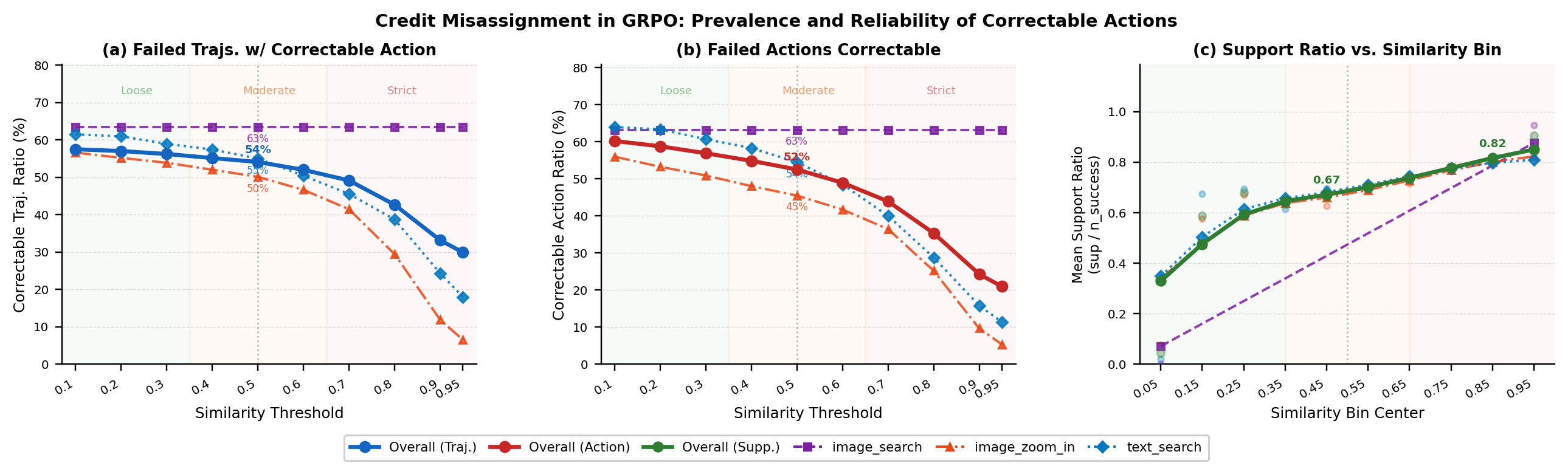}
  \caption{Prevalence and reliability of correctable credit misassignment in GRPO.
  A failing action is considered correctable if a parameter-equivalent action (similarity $\geq \theta$) exists in at least one successful trajectory on the same question.
  (a)~Fraction of failing trajectories containing at least one correctable tool-use action, as a function of similarity threshold $\theta$.
  (b)~Fraction of failing tool-use actions that are correctable.
  (c)~Mean support ratio ($\text{sup}/n_{\text{success}}$) per similarity bin, showing that higher-similarity matches consistently correspond to greater coverage across successful trajectories, confirming their reliability as correction signals.
  }
  \label{fig:prevalence}
\end{figure*}

\paragraph{Credit Misassignment: A Concrete Example.}
As illustrated in Figure~\ref{fig:example}, two trajectories share identical tool calls yet reach different final answers.
Under GRPO, the same Actions~1 and~2 receive positive advantages in the successful trajectory but negative advantages in the failing one---identical parameters, opposite gradient signals, within the same training step, yielding contradictory and self-canceling optimization.
We formalize this failure mode as \textbf{credit misassignment}: GRPO ignores the actual informational value of tool-use steps and uniformly assigns a trajectory-level negative advantage to all steps in any failing trajectory.

\paragraph{Scale of Credit Misassignment: Empirical Quantification.}
Credit misassignment is not a rare edge case---at a moderate threshold ($\theta \approx 0.5$), over half of failing trajectories and failing tool-use actions exhibit correctable credit misassignment (Figure~\ref{fig:prevalence}(a)(b)).
Moreover, parameter similarity and reference coverage are monotonically correlated (Figure~\ref{fig:prevalence}(c)): higher-similarity matches consistently correspond to greater coverage across successful trajectories, constituting reliable correction signals that actively filter out incidental noise from low-confidence matches.
These statistics confirm that the wasted training signal is both substantial and structurally exploitable.

A natural remedy is fine-grained credit assignment, but existing methods are all prohibitively costly~\cite{lightman2023let,wang2024math,luo2024improve,zhang2026reasoning}: manual annotation, dedicated PRM training, and Monte Carlo continuation sampling are expensive and none guarantees accurate credit assignment~\cite{zhang2026reasoning}.
Critically, all of them treat tool-use steps as ordinary generation steps, overlooking a structural property unique to tool calls that makes accurate credit assignment possible \emph{without any additional cost}.

\paragraph{Key Insight: Parameter-Deterministic Tools.}
We identify a key property of \textbf{information-acquisition tools}: \textbf{parameter determinism}. That is, the call parameters determine the underlying information-acquisition action, and similar parameters correspond to similar information-acquisition intent.
This property holds concretely for image search, region zoom-in, and text search, which respectively acquire search-image information, image-region information, and textual search information.
Crucially, it enables a conservative, batch-internal form of counterfactual reasoning: if a tool-use step in a failing trajectory is parameter-equivalent to steps in successful trajectories, those successes can serve as \textbf{counterfactual witnesses}, showing that the same information-acquisition action is useful and that its assigned negative advantage is a credit misassignment worth partially correcting.

\paragraph{Our Method: TAPO.}
Building on this insight, we propose \textbf{Tool-Aware Policy Optimization (TAPO)}, which constructs a counterfactual reference library from successful trajectories in the current batch, uses a confidence score (parameter similarity $\times$ coverage) to gate credit transfer, and applies a conservative clamp ensuring no failing step is globally encouraged---requiring no additional models, annotation, or sampling, and integrating as a drop-in replacement for the advantage assignment step of any group-based RL algorithm.
TAPO naturally achieves computational efficiency, difficulty adaptivity, and safe degradation to standard GRPO, and is instantiated on three tool types: image search, region zoom-in, and text search.

Our main contributions are as follows:
\begin{itemize}
  \item We identify and formally characterize \textbf{credit misassignment} as a systematic failure mode of GRPO in tool-augmented reasoning, and empirically quantify its prevalence and scale.
  \item We propose \textbf{TAPO}, a plug-and-play method that exploits parameter determinism to compensate misassigned negative credit at zero marginal cost, requiring no additional models, annotation, or sampling.
  \item TAPO consistently improves performance across multiple multimodal search benchmarks and three RL algorithms with negligible computational overhead.
\end{itemize}

\section{Related Work}

\paragraph{Multimodal Search Agents.}
To mitigate the limitations of static knowledge bases, large language models have evolved from static retrieval mechanisms into dynamic tool-augmented reasoning systems~\cite{chng2025sensenova}.
In the multimodal search domain, MMSearch-R1~\cite{wu2025mmsearch} and WebWatcher~\cite{geng2025webwatcher} build multimodal search agents by designing image search and text search tools.
DeepEyesV2~\cite{hong2025deepeyesv2} and SenseNova-MARS~\cite{chng2025sensenova} further introduce image processing tools to enable fine-grained multimodal search and understanding.
Vision-DeepSearch~\cite{zeng2026visiondeepresearchbenchmarkrethinkingvisual} and MTA-Agent~\cite{peng2026mtaagentopenrecipemultimodal} further advance agent performance through more sophisticated data construction and training pipelines.
However, these multimodal search methods focus exclusively on improving overall performance, overlooking the intrinsic credit assignment problem inherent in multi-step tool use.

\paragraph{Fine-Grained Credit Assignment in Agentic RL.}

PRM-based methods~\cite{lightman2023let, wang2024math} and step-level sampling approaches~\cite{luo2024improve, yue2026promoting} can provide step-wise reward signals, but at prohibitive cost and without guaranteeing accuracy. More recently, lighter-weight alternatives have emerged: GiGPO~\cite{feng2025groupingrouppolicyoptimizationllm} constructs step-level rewards via state grouping; HCAPO~\cite{tan2026hindsight} and Belief-RL~\cite{auzina2026intrinsiccreditassignmentlong} assess step quality through hindsight probability shifts and belief estimation, respectively. However, these methods predominantly target tool-integrated reasoning in the text domain. Critically, all of these methods treat tool-use steps as ordinary generation steps, overlooking the \emph{parameter-determinism} property of information-acquisition tools: similar call parameters define equivalent information-acquisition actions, making it possible to compare action credit across successful and failing trajectories.

\section{Method}

\begin{figure*}[t]
  \centering
  \includegraphics[width=\textwidth]{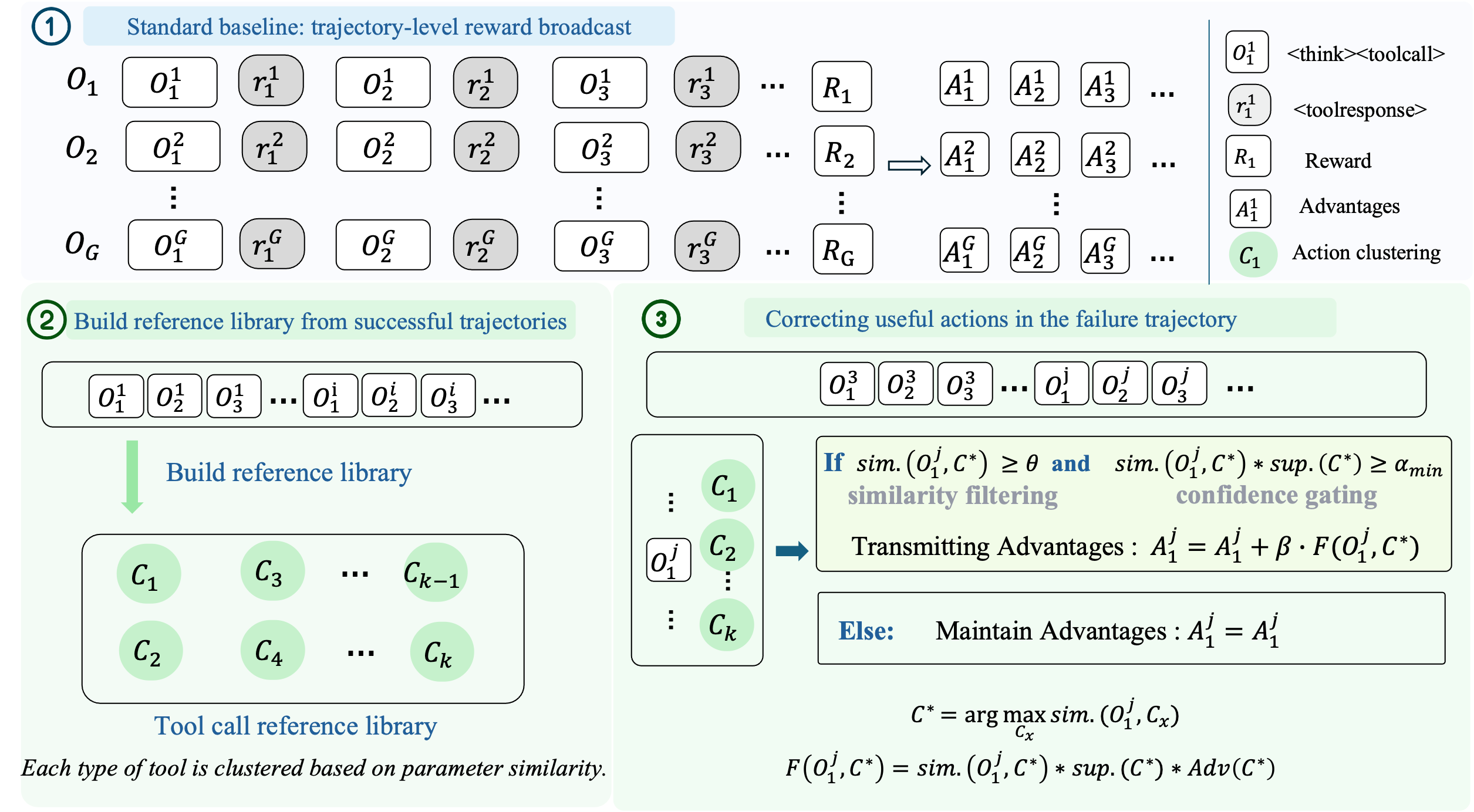}
  \caption{Overview of TAPO.
  \protect\ding{192}~Standard GRPO broadcasts trajectory-level advantages uniformly to all tokens.
  \protect\ding{193}~Tool-use steps from successful trajectories are clustered into reference groups.
  \protect\ding{194}~Each failing tool-use step is matched to the most similar reference group, and advantage transfer is applied via a two-level gate ($\theta$ and $\alpha_{\min}$).
  }

  \label{fig:pipeline}
\end{figure*}

\subsection{Problem Formulation}

\paragraph{Tool-augmented reasoning trajectories.}
Let a multimodal tool-augmented reasoning trajectory be:
$$\tau^j = \left( o_1^j,\ r_1^j,\ o_2^j,\ r_2^j,\ \ldots,\ o_K^j,\ r_K^j,\ o_\text{ans}^j \right)$$
where $o_k^j$ denotes the $k$-th reasoning and tool-use token segment, $r_k^j$ the corresponding tool-response segment, and $o_\text{ans}^j$ the final answer segment; $\text{args}(o_k^j)$ and $\text{type}(o_k^j)$ denote the call arguments and tool type of $o_k^j$, respectively.
The outcome reward $R_j$ is dominated by an accuracy component $r_{\text{acc}} \in \{0,1\}$ that depends solely on the correctness of the final answer. GRPO computes group-normalized advantages $\hat{A}_j = (R_j - \mu) / (\sigma + \varepsilon)$,
where $\mu$ and $\sigma$ are the mean and standard deviation of rewards within the group,
and broadcasts $\hat{A}_j$ to every response token in $\tau^j$; the policy gradient objective follows the standard PPO-clip form.

\subsection{Formal Analysis of Credit Misassignment}

\paragraph{Estimation bias of GRPO.}
For a failing trajectory $\tau^j$ ($r_{\text{acc}} = 0$, $\hat{A}_j < 0$), the true action value of tool-use step $o_k^j$ is:
$$Q^\pi(o_k^j) = \mathbb{E}_{\tau \sim \pi}\!\left[R \mid o_k^j \right]$$
Decomposing by trajectory outcome (law of total expectation):
\begin{multline}
  Q^\pi(o_k^j) =
  \underbrace{p^+_k \cdot \mathbb{E}\!\left[\hat{A} \mid o_k^j,\ \tau \text{ succeeds}\right]}_{\text{(I): term omitted by GRPO}} \\
  + \underbrace{(1-p^+_k) \cdot \mathbb{E}\!\left[\hat{A} \mid o_k^j,\ \tau \text{ fails}\right]}_{\text{(II): term used by GRPO}}
\end{multline}
where $p^+_k = P(\tau \text{ succeeds} \mid o_k^j)$.
GRPO assigns a uniform $\hat{A}_j < 0$ to every tool step in failing trajectories, effectively using only term~(II) while systematically ignoring term~(I).
When $p^+_k > 0$, the advantage estimation bias is:
$$\varepsilon_k \triangleq Q^\pi(o_k^j) - \hat{A}_j = p^+_k \cdot \mathbb{E}\!\left[\hat{A} \mid o_k^j,\ \tau^+\right] > 0$$

The objective of TAPO is to estimate and compensate the bias $\varepsilon_k$ for each failing tool-use step, without introducing any annotation or model.

\paragraph{Definition 1 (Credit Misassignment).}
An optimization method exhibits credit misassignment if a tool-use step $o_k^j$ in a failing trajectory has positive informational value (i.e., $p^+_k > 0$) yet is assigned the same trajectory-level negative advantage as tool uses with no informational value ($p^+_k \approx 0$).
Determining whether $p^+_k > 0$ requires parameter-level equivalence judgments, not trajectory-level rewards---which is precisely what the counterfactual witness construction below provides.

\subsection{Parameter-Deterministic Tools and Counterfactual Credit Transfer}

\paragraph{Assumption 1 (Parameter Determinism).}
For information-acquisition tools, if the parameters of two calls satisfy the similarity threshold
$\text{sim}(\text{args}(o_k^j),\, \text{args}(o_{k'}^i)) \geq \theta$,
the two calls correspond to equivalent \emph{information-acquisition actions}: even when the execution environment is noisy, they target the same evidence need and should therefore carry consistent action-level informational value.
What parameter determinism requires is that the useful information sought by the action is primarily determined by the call parameters, rather than by the surrounding trajectory context.
This assumption naturally extends to tools where fixed parameters determine the underlying information-acquisition operation,
and is concretely instantiated for the three tool types used in this work (see \S\ref{sec:appendix_impl} for details).

\paragraph{Counterfactual Witness.}
%

Under Assumption~1, if tool-use step $o_k^j$ in failing trajectory $\tau^j$ is parameter-equivalent to a step in successful trajectory $\tau^i$, then $\tau^i$ constitutes a \textbf{counterfactual witness}, showing that this information-acquisition action can make a positive contribution to solving the question and therefore deserves positive credit.
Accordingly, assigning the matched failing step the same uniform negative advantage as the rest of the failed trajectory constitutes credit misassignment warranting compensation.
Parameter equivalence judgments rely solely on tool arguments from trajectories already in the current batch, requiring no additional sampling, models, or annotation, and thus achieving approximate process supervision at zero marginal cost.

\subsection{Confidence-Gated Credit Transfer}

\paragraph{Reference library construction.}
For all successful trajectories on the same question,
tool-use steps of the same tool type with equivalent parameters are clustered into reference groups
$C_1, C_2, \ldots, C_k$ (Figure~\ref{fig:pipeline}~\ding{193}).
The reference advantage of each group is:
$$\text{Adv}(C_x) = \frac{1}{|C_x|}\sum_{o_{k'}^i \in C_x} \hat{A}_i$$
where GRPO-normalized advantages $\hat{A}$ (rather than raw rewards $R$) are aggregated.

\paragraph{Best-match selection.}
For a failing step $o_k^j$, the best-matching reference group is selected by maximum parameter similarity
(Figure~\ref{fig:pipeline}~\ding{194}):
\begin{align}
  C^* &= \arg\max_{C_x}\ \text{sim.}(o_k^j,\,C_x) \nonumber\\
  \text{sim.}(o_k^j, C_x) &= \frac{1}{|C_x|}\sum_{o_{k'}^i \in C_x}\text{sim}(o_k^j,\,o_{k'}^i)
\end{align}
where $\text{sim}(o_k^j, o_{k'}^i)$ measures the semantic similarity between the tool-call parameters of steps $o_k^j$ and $o_{k'}^i$.
If $\text{sim.}(o_k^j, C^*) < \theta$, no transfer is performed.

\paragraph{Credit transfer function.}
Given the selected $C^*$, the credit transfer function is:
\begin{equation}
  F(o_k^j, C^*) =
  \underbrace{\text{sim.}(o_k^j, C^*)}_{\text{param.\ match}} \times
  \underbrace{\text{sup.}(C^*)}_{\text{coverage}} \times \text{Adv}(C^*)
\end{equation}
where $\text{sup.}(C^*) = n(C^*)/G^+$, with $n(C^*)$ the number of successful trajectories containing a step in $C^*$ and $G^+$ the total number of successful trajectories.
The confidence score $\alpha = \text{sim.} \times \text{sup.}$ jointly captures parameter match quality and signal coverage: higher similarity indicates stronger action-credit equivalence, while higher support indicates a more robust reference signal with stronger positive credit; this design approximates the omitted term~(I) in GRPO's advantage estimation.
If $\alpha(o_k^j, C^*) < \alpha_{\min}$, no credit transfer is applied, implementing a two-level gate: $\theta$ first filters by parameter similarity; $\alpha_{\min}$ then gates on the joint confidence score, requiring both high match quality and sufficient coverage.

\subsection{Conservative Advantage Compensation}

%
%
For each failing tool-use step $o_k^j$ with $\alpha(o_k^j, C^*) \geq \alpha_{\min}$, the compensated advantage is:
\begin{equation}
  A_k^j \leftarrow \min\!\left(\hat{A}_j + \beta \cdot F(o_k^j, C^*),\ 0\right)
\end{equation}
where $\beta > 0$ is the transfer coefficient; otherwise $A_k^j \leftarrow \hat{A}_j$.
Clipping at $0$ prevents any tool step in a failing trajectory from receiving positive advantage, preserving the overall optimization direction.
TAPO therefore requires no modification to the optimizer or sampling procedure and serves as a drop-in replacement for GRPO's advantage assignment.
It also provides two benefits: difficulty adaptivity, since on hard questions ($G^+ \ll G$), rare successes yield larger normalized advantages and stronger credit transfer, while on easy questions ($\hat{A}_\text{success} \approx 0$), TAPO reduces to standard GRPO; and entropy preservation, since partially canceling unwarranted negative gradients on tool calls that deserve positive credit sustains exploration and mitigates entropy collapse, explaining the entropy trends observed across all three RL algorithms (Figure~\ref{fig:ablation_algo}).

\begin{table*}[t]
\setlength{\tabcolsep}{3pt}
\centering
\resizebox{\textwidth}{!}{%
\begin{tabular}{c|c|c|ccccccc}
\toprule
\textbf{Type} & \textbf{Model} & \textbf{Average} & \textbf{MMSearch} & \textbf{HR-MMSearch} & \textbf{FVQA-test} & \textbf{InfoSeek} & \textbf{SimpleVQA} & \textbf{LiveVQA} & \textbf{MAT-Search} \\
\midrule
\multicolumn{10}{c}{\textit{\textbf{Direct Answer}}} \\ \midrule
\multirow{3}{*}{\shortstack{Open-source}} 
 & Qwen2.5-VL-7B-Instruct \cite{bai2025qwen2}    & 27.70 & 7.60  & 0.58  & 26.28 & 31.95 & 47.88 & 19.63 & 60.00 \\
 & Qwen3-VL-8B-Instruct \cite{QwenTeam2025QwenVL}          & 29.24 & 11.70 & 12.13 & 24.22 & 23.15 & 42.94 & 23.18 & 67.33 \\
 & Qwen2.5-VL-32B-Instruct \cite{bai2025qwen2}  & 32.01 & 11.70 & 3.93  & 30.50 & 36.65 & 48.57 & 21.40 & 71.33 \\
 & Qwen3-VL-32B-Instruct \cite{QwenTeam2025QwenVL}  & 35.22 & 16.96 & 19.02  & 32.17 & 28.95 & 45.90 & 31.59 & 72.67 \\
 \midrule
\multirow{5}{*}{Proprietary} 
 & GPT-4o-mini \cite{hurst2024gpt}              & 33.08 & 15.79 & 1.31  & 36.83 & 35.95 & 44.42 & 24.63 & 72.66 \\
 & Gemini-2.5-Flash \cite{comanici2025gemini}   & 40.87 & 21.64 & 7.54  & 43.78 & 44.10 & 55.48 & 31.57 & 82.00 \\
 & GPT-4o \cite{hurst2024gpt}                    & 42.38 & 23.39 & 13.11 & 48.00 & 52.90 & 51.73 & 28.18 & 79.33 \\
 & GPT-5 \cite{openai2025gpt5}                  & 50.24 & 35.09 & 22.62 & 54.39 & 54.15 & 61.70 & 44.39 & 79.33 \\
  & GPT-5.2 \cite{openai2025gpt5}                  & 50.92 & 43.27 & 24.92 & 50.94 & 50.40 & 59.92 & 47.00 & 80.00 \\
 & Gemini-3-Flash \cite{gemini3flash}            & 53.68 & 57.31 & 21.97 & 56.50 & 54.85 & 63.57 & 38.90 & 82.67 \\
& Gemini-3-Pro \cite{gemini3pro}            & 55.87 & 62.57 & 26.89 & 59.22 & 56.30  & 64.07 & 40.06 & 82.00 \\
\midrule
\multicolumn{10}{c}{\textit{\textbf{Agentic Model (zero-shot)}}} \\ \midrule
\multirow{3}{*}{\shortstack{Open-source}} 
 & Qwen2.5-VL-7B-Instruct \cite{bai2025qwen2}    & 35.50 & 32.16 & 19.34 & 36.00 & 28.80 & 42.35 & 22.52 & 67.33 \\
& Qwen3-VL-8B-Instruct \cite{QwenTeam2025QwenVL}          & 51.33 & 47.37 & 27.87 & 53.61 & 46.15 & 62.29 & 39.37 & 82.67 \\ 
 & Qwen2.5-VL-32B-Instruct \cite{bai2025qwen2}  & 53.45 & 49.71 & 33.44 & 52.22 & 50.10 & 65.15 & 42.17 & 81.33 \\
 & Qwen3-VL-32B-Instruct \cite{QwenTeam2025QwenVL}  & 53.82 & 49.12 & 34.43 & 54.28 & 49.85 & 64.17 & 42.87 & 82.00 \\ \midrule
\multirow{5}{*}{Proprietary} 
 & GPT-4o-mini \cite{hurst2024gpt}              & 45.65 & 38.60 & 26.23 & 50.00 & 42.35 & 50.84 & 31.54 & 80.00 \\
  & GPT-4o \cite{hurst2024gpt}                    & 55.09 & 49.12 & 30.16 & 66.34 & 59.55 & 63.67 & 40.09 & 76.67 \\
 & Gemini-2.5-Flash \cite{comanici2025gemini}   & 58.05 & 59.06 & 40.00 & 61.72 & 53.70 & 68.81 & 47.75 & 75.33 \\
 & GPT-5 \cite{openai2025gpt5}                  & 60.12 & 52.63 & 38.36 & 62.61 & 55.95 & 70.58 & 56.02 & 84.67 \\
 & Gemini-3-Flash \cite{gemini3flash}            & 61.26 & 62.57 & 41.64 & 64.89 & 61.10 & 67.92 & 48.06 & 82.67 \\
  & GPT-5.2 \cite{openai2025gpt5}                 & 67.64 & 66.08 & 48.20 & 68.78 & 65.55 & 78.18 & 65.99 & 80.67\\
 & Gemini-3-Pro \cite{gemini3pro}            & 69.06 & 74.27 & 48.52 & 72.61 & 66.45 & 75.91 & 59.69 & 86.00 \\
\midrule
\multicolumn{10}{c}{\textit{\textbf{Agentic Model}}} \\ \midrule
\multirow{6}{*}{\shortstack{Open-source}} 
 & Visual-ARFT \cite{liu2025visual}             & 40.13 & 34.50 & 24.92 & 41.72 & 37.95 & 42.45 & 25.40 & 74.00 \\
 & DeepMMSearch-R1 \cite{narayan2025deepmmsearch} & --    & --    & --    & --    & 47.51 & 55.87 & --    & -- \\
 & MMSearch-R1 \cite{wu2025mmsearch}             & 52.49 & 53.80 & 20.33 & 58.40 & 55.10 & 57.40 & 48.40 & 74.00 \\
 & DeepEyesV2 \cite{hong2025deepeyesv2}          & --    & 63.70 & --    & 60.60 & 51.10 & 59.40 & --    & -- \\
 & SkyWork-R1V4~\cite{zhang2025skyworkr1v4agenticmultimodalintelligence}  & -- & 66.10 & -- & 67.20 & -- & -- & -- & -- \\
 & SenseNova-MARS~\cite{chng2025sensenova}         & 63.65 & 66.67 & 40.33 & 67.11 & 61.70 & 70.19 & 56.22 & \textbf{83.33} \\
 & Qwen3-VL-8B-Instruct+SAPO   & 62.48 & 64.33 & 38.69 & 65.61 & 59.55 & 70.38 & 56.86 & 82.00 \\
 & TAPO (applied in SAPO)  & \textbf{65.21} & \textbf{67.21} & \textbf{40.66} & \textbf{69.89} & \textbf{62.70} & \textbf{73.64} & \textbf{59.71} & 82.67 \\

\bottomrule
\end{tabular}%
}
\caption{Performance on search-oriented benchmarks under Direct Answer and Agentic Model workflows.}
\label{tab:agentic_search}
\end{table*}

\begin{table*}[t]
\centering
\small
\begin{tabular}{lcccccccc}
\toprule
Method & MMSearch & HR-MMSearch & FVQA-test & SimpleVQA & LiveVQA & MAT-Search & InfoSeek & AVG \\
\midrule
GRPO                 & 53.22 & 29.51 & 62.17 & 67.11 & 51.42 & 78.00 & 52.45 & 56.27 \\
+TAPO  & 54.97 & 33.11 & 65.56 & 68.31 & 54.05 & 82.00 & 58.85 & 59.55 \\
$\Delta$             & +3.30\% & +12.21\% & +5.46\% & +1.78\% & +5.12\% & +5.13\% & +12.20\% & +5.83\% \\
\midrule
GSPO                 & 53.22 & 31.15 & 66.50 & 68.21 & 53.55 & 80.67 & 56.40 & 58.53 \\
+TAPO  &60.82 & 31.48 & 66.61 & 70.98 & 56.14 & 83.33 & 59.40 & 61.25 \\
$\Delta$             & +14.28\% & +1.05\% & +0.17\% & +4.06\% & +4.82\% & +3.31\% & +5.32\% & +4.65\% \\
\midrule
SAPO                 & 53.97 & 30.16 & 64.50 & 67.52 & 51.75 & 81.33 & 55.10 & 57.76 \\
+TAPO  & 56.73 & 31.48 & 67.17 & 71.27 & 55.36 & 81.33 & 59.95 & 60.47 \\
$\Delta$             & +5.10\% & +4.35\% & +4.13\% & +5.56\% & +6.97\% & +0.00\% & +8.80\% & +4.69\% \\
\bottomrule
\end{tabular}
\caption{Ablation on the base algorithm for TAPO ($\beta$=0.25).
All values are accuracy (\%) on seven multimodal search benchmarks.
$\Delta$ denotes the relative improvement (\%) of TAPO over the corresponding base algorithm.}
\label{tab:ablation_algo}
\end{table*}


\section{Experiments}

\subsection{Implementation Details}

\paragraph{Tool Setup.}
Following the setup of SenseNova-MARS~\cite{chng2025sensenova}, we instantiate three representative tool types:
(1) \textit{Image search} performs reverse image search and returns image captions;
(2) \textit{Text search} executes a query-specified web search and returns a Qwen3-32B-generated summary;
(3) \textit{Region zoom-in} crops a specified bounding-box region and returns the patch.
Parameter similarity definitions for each tool type and a per-type training analysis are provided in Appendix~\ref{sec:appendix_param_sim} and~\ref{sec:appendix_per_tool}.

\paragraph{Training.}
We train with the VeRL framework~\cite{sheng2024hybridflow} without any SFT warm-up, using a global batch size of 128, a learning rate of $1 \times 10^{-6}$, and $G=8$ rollout trajectories per question on approximately 4K multimodal search samples.
The reward function combines an accuracy reward and a format reward (see Appendix~\ref{sec:appendix_reward} for details).
To reduce training cost, text search during training is implemented via a local Wikipedia-based RAG system; full training configuration is provided in Appendix~\ref{sec:appendix_data}.


\paragraph{Evaluation.}
We evaluate on seven multimodal search benchmarks (MMSearch, HR-MMSearch, FVQA-test, InfoSeek, SimpleVQA, LiveVQA, MAT-Search) using the Serper API; benchmark details are in Appendix~\ref{sec:appendix_benchmarks}.
We compare against top closed-source models (GPT-5, Gemini~3, in both direct-answer and agentic modes), leading open-source models (Qwen3-VL), and competitive search agents (DeepEyesV2~\cite{hong2025deepeyesv2}, WebWatcher~\cite{geng2025webwatcher}, SenseNova-MARS~\cite{chng2025sensenova}).


\begin{table*}[htbp]
\centering
\resizebox{\textwidth}{!}{%
\begin{tabular}{lcccccccc}
\toprule
$\beta$ &	MMSearch	&	HR-MMSearch	&	FVQA-test	&	SimpleVQA	&	LiveVQA	&	MAT-Search	&	InfoSeek	&	AVG \\
\midrule
Qwen3-VL-4B & 43.86 & 25.57 & 50.50 & 63.57 & 37.34 & 79.33 & 43.25 & 49.06 \\
GRPO       & 53.22 & 29.51 & 62.17 & 67.11 & 51.42 & 78.00 & 52.45 & 56.27 \\
\midrule
TAPO $\beta$=0.1  & 54.97 & 30.49 & 66.00 & 69.79 & 53.66 & 78.67 & 60.05 & 59.09 \\
TAPO $\beta$=0.25 & 54.97 & 33.11 & 65.56 & 68.31 & 54.05 & 82.00 & 58.85 & 59.55 \\
TAPO $\beta$=0.4  & 58.48 & 33.44 & 64.83 & 69.89 & 54.47 & 84.00 & 59.05 & 60.59 \\
TAPO $\beta$=0.6  & 54.97 & 34.10 & 68.28 & 71.17 & 55.89 & 80.00 & 61.70 & 60.87 \\
TAPO $\beta$=0.8  & 61.41 & 33.11 & 66.44 & 69.10 & 56.30 & 84.67 & 57.25 & 61.18 \\
\midrule
TAPO $\beta$=1.0  & 57.31 & 30.49 & 66.00 & 69.79 & 53.66 & 80.67 & 60.05 & 59.71 \\
TAPO $\beta$=1.0 w/o sup. &52.63	&20.66	&59.22	&67.72	&51.83	&81.33 &51.60		&55.00 \\
TAPO $\beta$=1.0 w/o clip.&56.73	&29.84	&65.00	&69.10	&54.50  &80.00 &58.30	&59.07 \\
\midrule
$\Delta$ vs.\ Qwen3-VL-4B & +40.01\% & +33.33\% & +35.20\% & +11.96\% & +50.78\% & +6.73\% & +42.66\% & +24.71\% \\
$\Delta$ vs.\ GRPO & +15.40\% & +15.56\% & +9.83\%  & +6.05\%  & +9.50\%  & +8.55\% & +17.64\% & +8.74\%  \\
\bottomrule
\end{tabular}%
}%
\caption{Ablation on the transfer coefficient $\beta$ (TAPO applied to GRPO backbone).
All values are accuracy (\%) on seven multimodal search benchmarks.
$\Delta$~vs.\ Qwen3-VL-4B: improvement of the best-$\beta$ result over the Qwen3-VL-4B base model;
$\Delta$~vs.\ GRPO: improvement over vanilla GRPO training.}
\label{tab:ablation_beta}
\end{table*}

\begin{figure}[t]
  \centering
  \includegraphics[width=\linewidth]{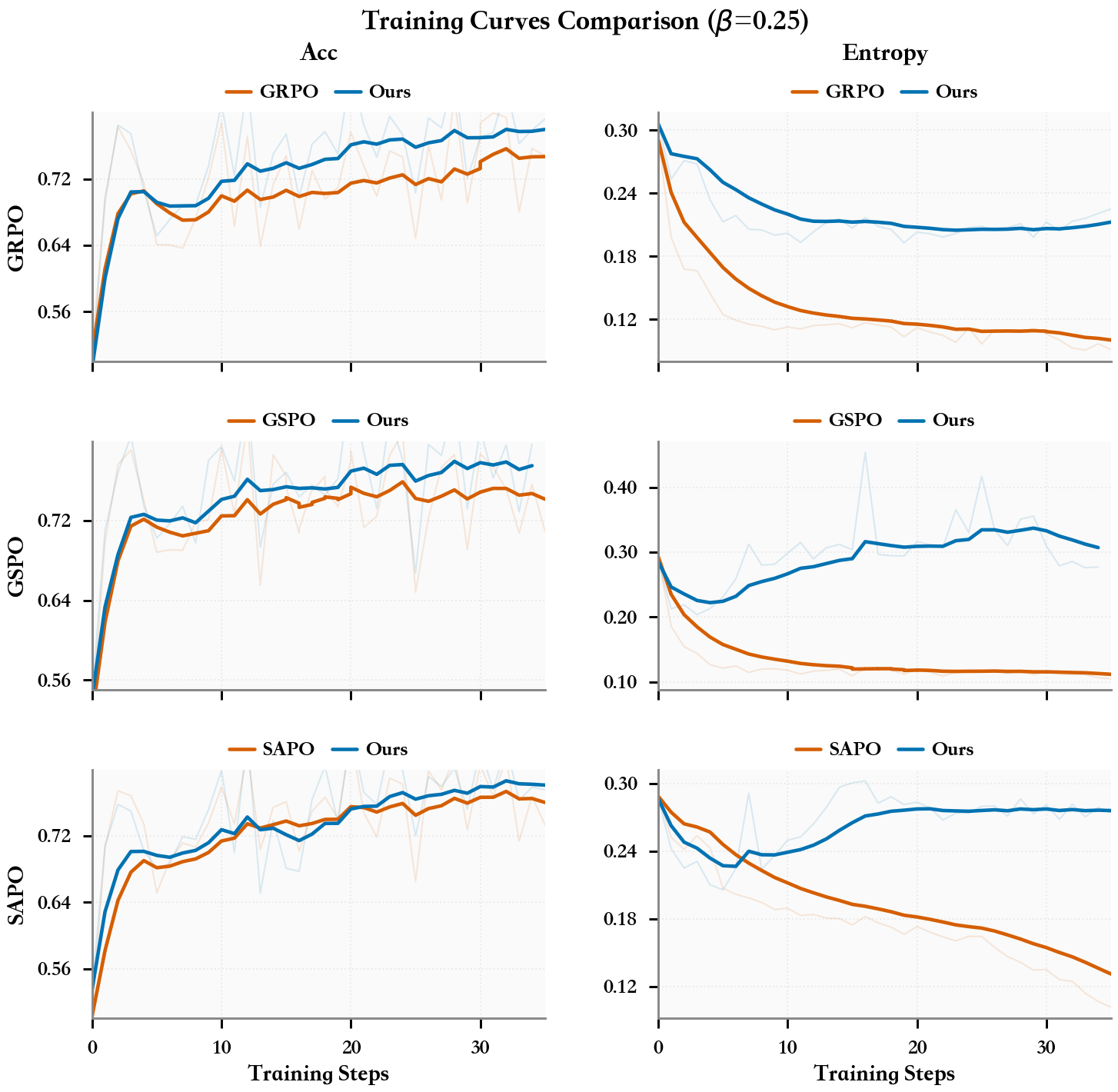}
\caption{Ablation on generality across RL algorithms.
}
  \label{fig:ablation_algo}
\end{figure}

\subsection{Main Results}
As shown in Table~\ref{tab:agentic_search}, TAPO achieves consistent improvements across all multimodal search benchmarks.
We adopt SAPO as our RL backbone, as it represents the strongest group-based RL baseline, lifting the base model Qwen3-VL-8B from 51.33\% to 62.64\% average accuracy. TAPO further improves upon this strong baseline by \textbf{4.4\%} relatively, demonstrating the additional headroom unlocked by correcting credit misassignment as an orthogonal optimization atop RL training.
Compared to the strong search agent baseline SenseNova-MARS, which shares the same data and tool configuration, TAPO further achieves a \textbf{2.5\%} relative improvement.
Notably, these gains stem entirely from TAPO's training algorithm, requiring no higher-quality data, additional tools, or increased model capacity. Moreover, the computational overhead is minimal: building the reference library and propagating advantages accounts for only \textbf{0.06\%} of total training time (under 1\,second per step). This highlights the effectiveness of correcting credit misassignment as an orthogonal, lightweight optimization that integrates seamlessly into any group-based RL algorithm.

\begin{figure*}[t]
  \centering
  \includegraphics[width=\textwidth]{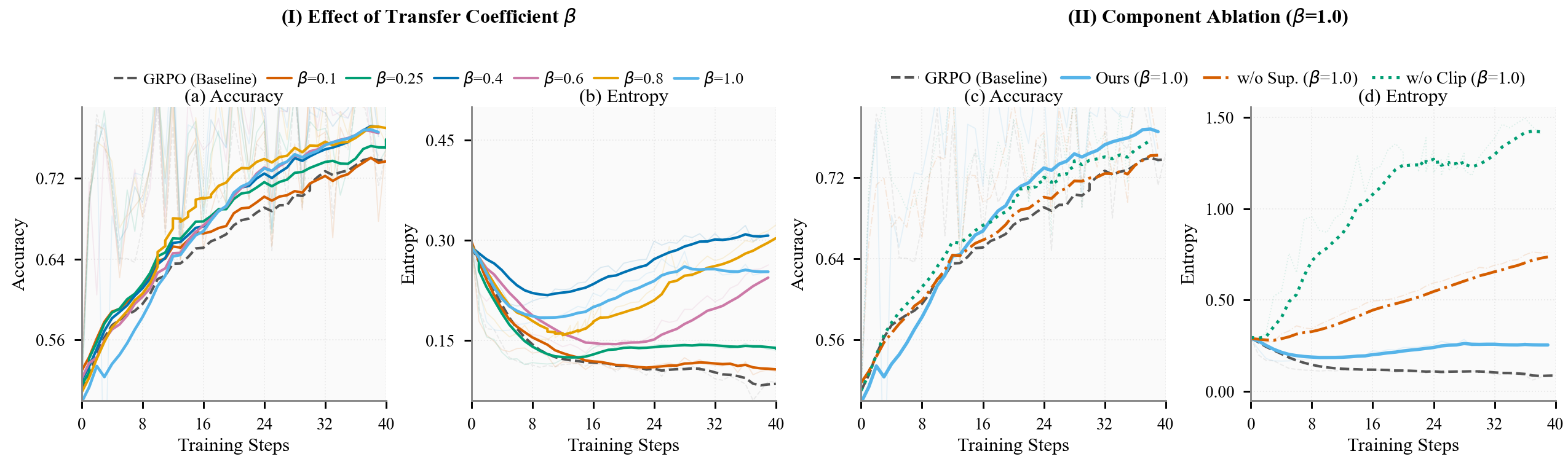}
  \caption{Robustness of TAPO to the transfer coefficient $\beta$.}
  \label{fig:ablation_beta}
\end{figure*}

\subsection{Ablation Study}
\paragraph{Generalization across RL algorithms.}
To verify that TAPO is algorithm-agnostic, we apply it to three representative RL algorithms: GRPO, GSPO, and SAPO.
All experiments use Qwen3-VL-4B for one training epoch, with the transfer coefficient fixed at $\beta = 0.25$.
Training curves and evaluation results are shown in Figure~\ref{fig:ablation_algo} and Table~\ref{tab:ablation_algo}.
As shown in the upper row of Figure~\ref{fig:ablation_algo}, TAPO consistently improves training accuracy over the baselines for all three algorithms.
The lower row further shows that TAPO mitigates entropy collapse across all three algorithms, preserving exploration.
Quantitatively, TAPO improves average accuracy by 4--6 percentage points across the three algorithms (Table~\ref{tab:ablation_algo}),
confirming its generality as a plug-and-play component.

\paragraph{Robustness to the transfer coefficient $\beta$.}

We sweep $\beta \in \{0.1, 0.25, 0.4, 0.6, 0.8, 1.0\}$ to test TAPO's sensitivity. Results are shown in Figure~\ref{fig:ablation_beta} and Table~\ref{tab:ablation_beta}; detailed dynamics for $\beta{=}0.25$ are deferred to Appendix~\ref{sec:appendix_tapo_dynamics}.
At small $\beta$ (e.g., $0.1$), transfer is weak and performance is close to GRPO (+2.82\% AVG). As $\beta$ increases, TAPO improves and peaks at $\beta = 0.8$. TAPO outperforms GRPO across all tested values, indicating robustness to $\beta$.

We further ablate the two key components of TAPO at $\beta = 1.0$: removing the support factor (\textit{w/o sup.}) and removing the conservative clamp (\textit{w/o clip.}).
Removing support degrades AVG from 59.71\% to 55.00\% ($-$4.71\%), confirming that the support factor is essential for filtering unreliable signals from small reference groups.
Removing the clamp causes notable entropy divergence during training (Figure~\ref{fig:ablation_beta}, right), as some tool-use steps in failing trajectories receive positive advantages, destabilizing the policy update; the final AVG (59.07\%) also falls below the full model, confirming that the upper-bound constraint is necessary for both training stability and final performance.

\subsection{Training Dynamics Analysis}
\label{sec:analysis}

\begin{figure}[htbp]
  \centering
  \includegraphics[width=\linewidth]{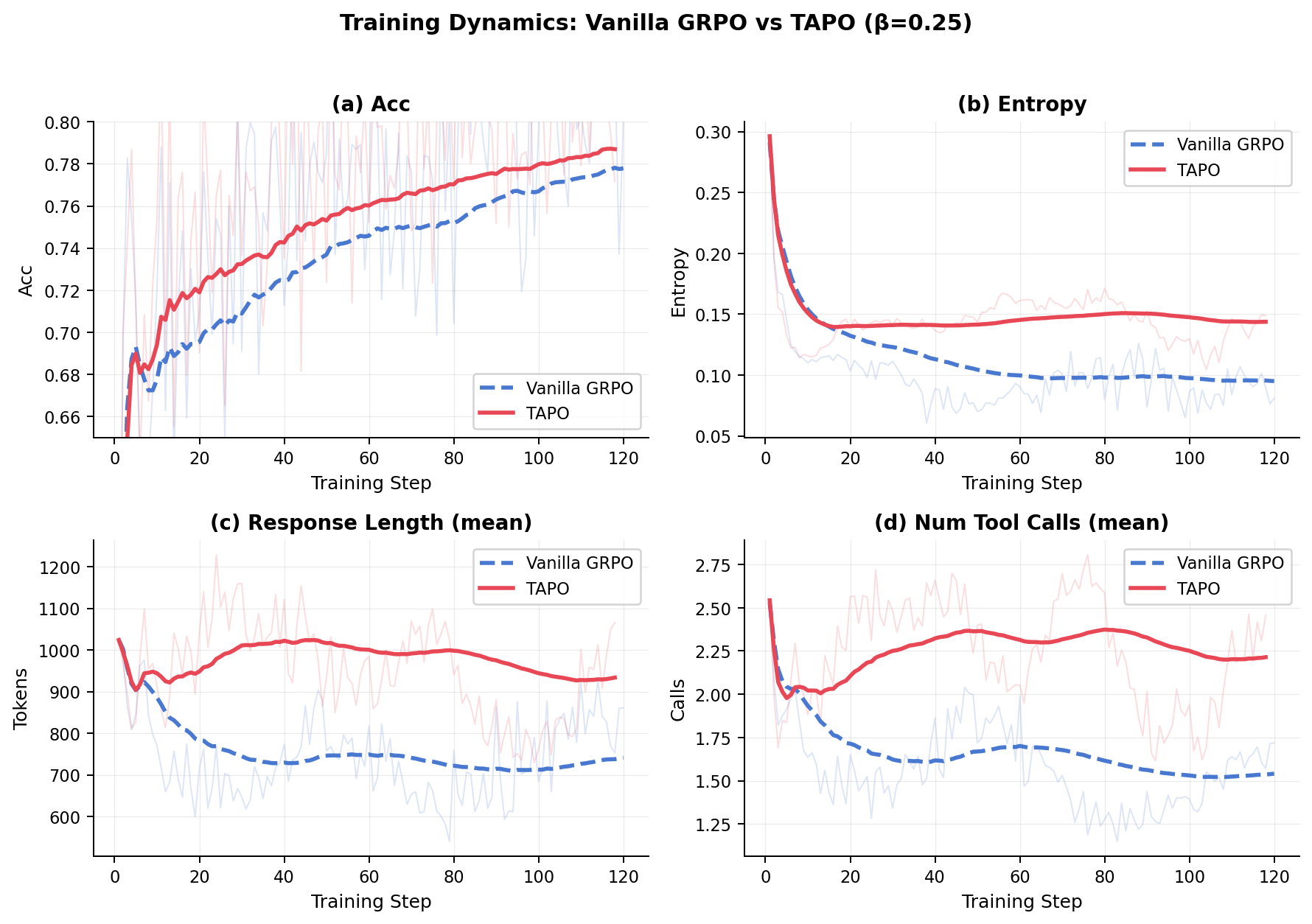}
  \caption{Training dynamics of Vanilla GRPO vs.\ TAPO ($\beta{=}0.25$) on Qwen3-VL-4B.
  }
  \label{fig:training_dynamics}
\end{figure}

The advantage compensation for credit misassignment not only improves final benchmark performance, but also induces observable behavioral differences throughout training (see Appendix~\ref{sec:appendix_misassign_persist} for an analysis of why credit misassignment persists structurally throughout training).
We compare the training dynamics of Vanilla GRPO and TAPO ($\beta{=}0.25$) along four dimensions: training accuracy, policy entropy, response length, and number of tool calls per rollout (Figure~\ref{fig:training_dynamics}).

\paragraph{Training accuracy and entropy.}
As shown in Figure~\ref{fig:training_dynamics}(a), TAPO consistently achieves higher training accuracy than Vanilla GRPO throughout the entire training process.
Figure~\ref{fig:training_dynamics}(b) and Figure~\ref{fig:ablation_beta}(b) reveal the underlying mechanism: Vanilla GRPO suffers from progressive entropy collapse, with policy entropy declining steadily, reflecting a deteriorating exploration capacity.
In contrast, by correcting the misassigned negative gradients on high-quality tool-use steps, TAPO provides more positive advantage encouragement compared to the vanilla algorithm, enhancing the model's beneficial exploration behavior and preventing the policy from prematurely converging to a degenerate, low-diversity behavior.

\paragraph{Response length and tool usage.}
Figures~\ref{fig:training_dynamics}(c) and~(d) reveal a consistent behavioral degradation under standard GRPO: both response length and tool call frequency decline progressively throughout training.
This is a direct consequence of credit misassignment---the model learns to reduce tool invocations as a shortcut to avoid negative gradient signals, sacrificing information-gathering depth in favor of shorter, less exploratory responses.
TAPO effectively reverses this degradation, with both response length and tool call frequency exhibiting more diverse and dynamic patterns throughout training.
These dynamics collectively demonstrate that credit misassignment persistently suppresses tool-use behavior during training, and TAPO effectively mitigates this negative effect through advantage compensation.

\section{Conclusion}

We identify and quantify credit misassignment as a systematic failure mode of GRPO in multimodal search, and propose TAPO as a lightweight remedy.
By exploiting the parameter-determinism property of information-acquisition tools, TAPO compensates misassigned advantages through confidence-gated credit transfer, requiring no additional annotation, models, or sampling while introducing only negligible computational overhead.
Across multiple multimodal search benchmarks, TAPO consistently improves over strong baselines, remains robust to $\beta$, is algorithm-agnostic, and effectively mitigates entropy collapse and tool-use degradation during training.
TAPO is orthogonal to existing multimodal search agent designs and can be seamlessly integrated into any group-based RL training pipeline built on parameter-deterministic information-acquisition tools.
More broadly, the parameter-determinism assumption may extend beyond the three validated tools to other tool-augmented settings where fixed parameters determine information-acquisition actions.




\bibliography{custom}

\appendix

\section{Multimodal Search Agents}
\label{sec:appendix_background}
Reinforcement learning has emerged as the dominant paradigm for training LLM-based agents, achieving significant advances across diverse domains including mathematical reasoning~\cite{shao2024deepseekmath,yue2026promoting}, code generation~\cite{jiang2025coderl}, and web navigation~\cite{wei2025webagent}. In tool-augmented settings, agents follow a think--act--observe loop~\cite{yao2022react}, progressively collecting information through multi-step interactions with external environments and achieving consistent performance gains across diverse application scenarios~\cite{dong2025training,qi2026eviagent,wei2025webagent}. With outcome correctness as the reward signal, Agentic RL paradigm enables efficient end-to-end training, supported by dedicated agentic RL frameworks~\cite{sheng2024hybridflow,chai2025rlfactory}.
Multimodal search is a representative scenario of tool-augmented reasoning, requiring models to answer open-domain questions over mixed image--text inputs. In this setting, agents built on multimodal foundation models such as Qwen2.5-VL~\cite{bai2025qwen2} invoke multiple rounds of image search and text retrieval in coordination, progressively integrating multimodal tool outputs into the reasoning chain to produce the final answer, achieving substantial performance gains across multiple open-domain visual question answering benchmarks~\cite{hong2025deepeyesv2,geng2025webwatcher,chng2025sensenova}.
In this work, we instantiate TAPO within the multimodal search setting using three tool types: image search , text search, and region zoom-in, serving visual entity grounding, background knowledge retrieval, and fine-grained visual understanding, respectively.

\section{Implementation Details}
\label{sec:appendix_impl}


\subsection{Training Data}
\label{sec:appendix_data}

RL training uses the same training set as SenseNova-MARS~\cite{chng2025sensenova}, comprising approximately 3,859 samples across four subsets:
\begin{table*}[h]
\centering
\small
\begin{tabular}{lcc}
\toprule
\textbf{Dataset} & \textbf{\#Samples} & \textbf{Reward} \\
\midrule
FVQA-train           & 1,894 & LLM-judge + format \\
DeepEyes-4K (non-MCQ)& 425   & LLM-judge + format \\
Visual-Probe-train   & 912   & LLM-judge + format \\
DeepEyes-4K (MCQ)    & 628   & Exact match + format \\
\bottomrule
\end{tabular}
\caption{Composition of the RL training set.}
\label{tab:training_data}
\end{table*}

All subsets are used exclusively for RL training.
No supervised fine-tuning (SFT) warm-up stage is applied; we train directly from the instruction-tuned Qwen3-VL-8B-Instruct checkpoint.
Training uses the VeRL framework~\cite{sheng2024hybridflow} with a global batch size of 128, a learning rate of $1\times10^{-6}$, and $G=8$ rollout trajectories per question.
Each trajectory allows up to 10 interaction rounds, with a per-round token budget of 8,192 and a cumulative trajectory limit of 32,768 tokens.
During training, text search is implemented via a local Wikipedia-based RAG system to reduce computational cost.

\subsection{Reward Function}
\label{sec:appendix_reward}

The outcome reward combines two components:

\paragraph{Accuracy reward.}
For non-MCQ samples, we use Qwen3-VL-32B-Instruct as an LLM-as-a-Judge (greedy decoding, temperature = 0): if the model's final answer is judged semantically correct, $r_{\text{acc}} = 1.0$; otherwise $r_{\text{acc}} = 0.0$.
For MCQ samples, rule-based exact match is applied, assigning $r_{\text{acc}} = 1.0$ for a correct choice and $0.0$ otherwise.

\paragraph{Format reward.}
At each non-terminal turn, the response must contain a \texttt{<thinking>} block followed by a single valid \texttt{<tool\_call>} in JSON Schema format.
At the terminal turn, the response must contain a \texttt{<thinking>} block followed by an \texttt{<answer>} tag.
Full compliance yields $r_{\text{format}} = 0.5$; any violation yields $0.0$.

The final outcome reward is $R = r_{\text{acc}} + r_{\text{format}} \in \{0,\, 0.5,\, 1.0,\, 1.5\}$.

\subsection{Parameter Similarity Computation}
\label{sec:appendix_param_sim}

TAPO defines parameter similarity separately for each tool type.
All similarity functions are rule-based and require no embedding models or external LLMs.

\paragraph{Image search (\texttt{image\_search\_tool}).}
This tool takes no text parameters; all image search calls for a given question share the same input image.
Parameter equivalence is therefore trivial: within each training batch, all image search calls for the same question are merged into a single reference group, and parameter similarity is identically $1.0$.
Since image search has no free parameters, the confidence score $\alpha$ is determined solely by the support factor.
We set $\alpha_{\min,\text{isearch}} = 1.0$, meaning advantage transfer is triggered only when the reference group covers all successful trajectories in the batch (i.e., support factor $= 1.0$).

\paragraph{Region zoom-in (\texttt{image\_zoom\_in\_tool}).}
Parameter equivalence is measured by bounding-box IoU.
During reference library construction (Phase 3), zoom-in calls with IoU $\geq \theta_{\text{IoU}} = 0.7$ are merged into the same reference group.
During advantage transfer (Phase 4), the minimum confidence threshold is $\alpha_{\min,\text{zoom}} = 0.5$.

\paragraph{Text search (\texttt{text\_search\_tool}).}
Parameter equivalence is measured by query string similarity, computed as a weighted combination of three lexical metrics:
\begin{equation}
\text{sim}(q_1, q_2) = 0.3 \cdot J + 0.5 \cdot \text{OC} + 0.2 \cdot \text{BG}
\end{equation}
where $J$ is the Jaccard coefficient over stopword-filtered tokens, $\text{OC}$ is the Overlap Coefficient (robust to asymmetric query lengths), and $\text{BG}$ is character-bigram Jaccard similarity (capturing morphological variants).
Reference library construction uses aggregation threshold $\theta_{\text{text}} = 0.8$; Phase 4 matching uses $\alpha_{\min,\text{text}} = 0.5$.

\subsection{Evaluation Benchmarks}
\label{sec:appendix_benchmarks}

We evaluate on seven publicly available multimodal search benchmarks that collectively span a broad range of retrieval difficulty, temporal freshness, and image resolution requirements.

\textbf{MMSearch}~\cite{jiang2024mmsearch} contains 300 image-accompanied questions drawn from news events and domain knowledge published after August 2024, testing the model's ability to ground visual observations in up-to-date factual knowledge. We follow the standard split of 171 questions used for system-level evaluation.

\textbf{HR-MMSearch}~\cite{chng2025sensenova} is a knowledge-intensive benchmark comprising 305 questions paired with high-resolution (4K) images sourced from international news agencies (2025 events), spanning eight topical domains. Approximately 60\% of questions are classified as hard, requiring three or more tool invocations to answer correctly.

\textbf{FVQA-test}~\cite{wu2025mmsearch} is a curated 1,800-sample test set consisting of three equal-sized subsets: FVQA-auto-vc (600 samples with automatic validation), InfoSeek Human Split (600 samples with manually corrected annotations), and an expert-annotated subset (600 samples).

\textbf{InfoSeek}~\cite{chen2023can} is a visual entity retrieval benchmark grounded in Wikidata, covering fine-grained factual attributes that are unlikely to be memorized during pre-training. We sample 2,000 instances from the official test split.

\textbf{SimpleVQA}~\cite{cheng2025simplevqa} focuses on factual accuracy, combining post-2023 VQA instances with expert-curated image-question pairs across 1,013 English samples.

\textbf{LiveVQA}~\cite{fu2025livevqa} aggregates 3,602 image-question pairs from major news outlets (CNN, BBC, etc.), spanning 14 thematic categories, and emphasizes temporal reasoning over recently published content.

\textbf{MAT-Search}~\cite{liu2025visual} is a manually crafted and human-verified benchmark of 150 examples designed to evaluate agentic multimodal reasoning, requiring models to handle composite multi-hop queries, retrieve external information, and invoke tools effectively across varying reasoning depths.

All benchmarks are evaluated with pass@1 accuracy; web search is powered by the Serper API.
For open-ended questions, Qwen3-VL-32B-Instruct serves as an LLM-as-a-Judge with greedy decoding (temperature~$= 0$).
For all benchmarks, we report the mean pass@1 accuracy over three independent runs.

\subsection{Computational Overhead of TAPO}
\label{sec:appendix_overhead}

\begin{figure}[t]
  \centering
  \includegraphics[width=0.75\linewidth]{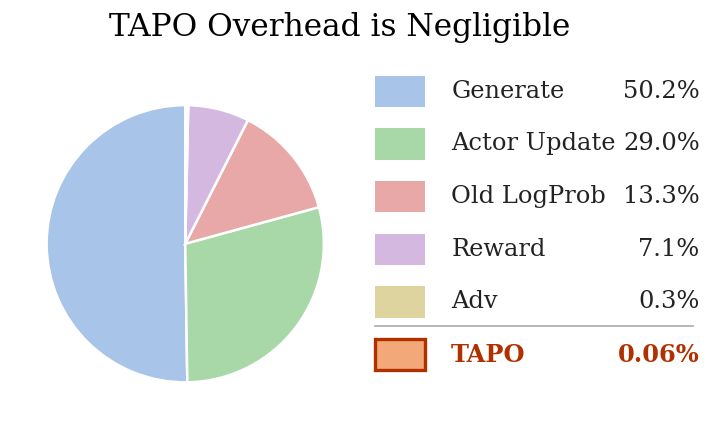}
  \caption{Wall-clock time breakdown per training step. TAPO (Phase~3+4) accounts for only \textbf{0.06\%}, negligible compared to the dominant components of token generation (50.2\%) and actor update (29.0\%).}
  \label{fig:timing_overhead}
\end{figure}

TAPO's Phase~3 (reference library construction) and Phase~4 (advantage transfer) run entirely on CPU using vectorized NumPy operations and lightweight string matching---no neural network forward passes are required.
Under our training configuration (global batch size 128, $G=8$ rollouts per question), these two phases together take under 1 second per training step, accounting for only \textbf{0.06\%} of total wall-clock training time, as illustrated in Figure~\ref{fig:timing_overhead}.
TAPO therefore introduces negligible computational overhead and can be seamlessly integrated into any group-based advantage estimation algorithm.

\section{Detailed Analysis of TAPO}
\label{sec:appendix_tapo_analysis}

\subsection{Persistence of Credit Misassignment During Training}
\label{sec:appendix_misassign_persist}

\begin{figure}[t]
  \centering
  \includegraphics[width=\linewidth]{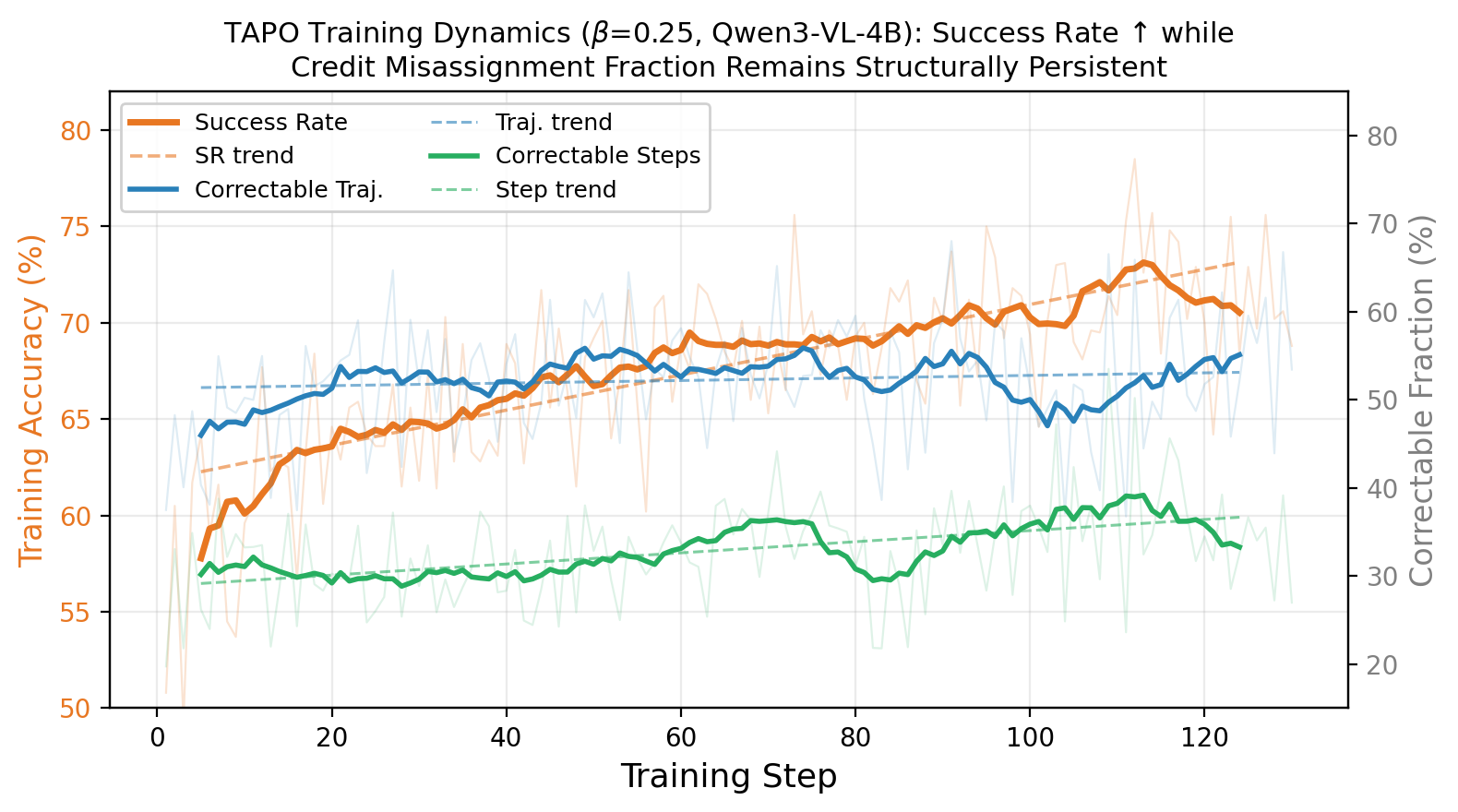}
  \caption{Training dynamics of TAPO ($\beta{=}0.25$, Qwen3-VL-4B).
  Training accuracy increases steadily throughout training, while the fraction of correctable trajectories (Correctable Traj.)\ and correctable steps (Correctable Steps) remain structurally stable and exhibit a slow upward trend, indicating that credit misassignment persists across the entire training horizon.}
  \label{fig:appendix_misassign_persist}
\end{figure}

As shown in Figure~\ref{fig:appendix_misassign_persist}, TAPO training leads to a continuous improvement in training accuracy, yet the fractions of correctable trajectories and correctable steps do not decline in tandem; instead, they remain structurally stable throughout training.
This indicates that in multi-tool search tasks, the phenomenon of high-quality tool-use steps appearing within failing trajectories remains prevalent as training progresses.
As the model learns more complex multi-step reasoning strategies, such ``locally effective but globally failing'' trajectories do not disappear naturally with improving model capability.
Consequently, TAPO's advantage compensation mechanism consistently faces meaningful correction targets throughout the entire training cycle, rather than being effective only in the early stages.

\subsection{Detailed Training Dynamics of TAPO }
\label{sec:appendix_tapo_dynamics}

\begin{figure*}[t]
  \centering
  \includegraphics[width=\textwidth]{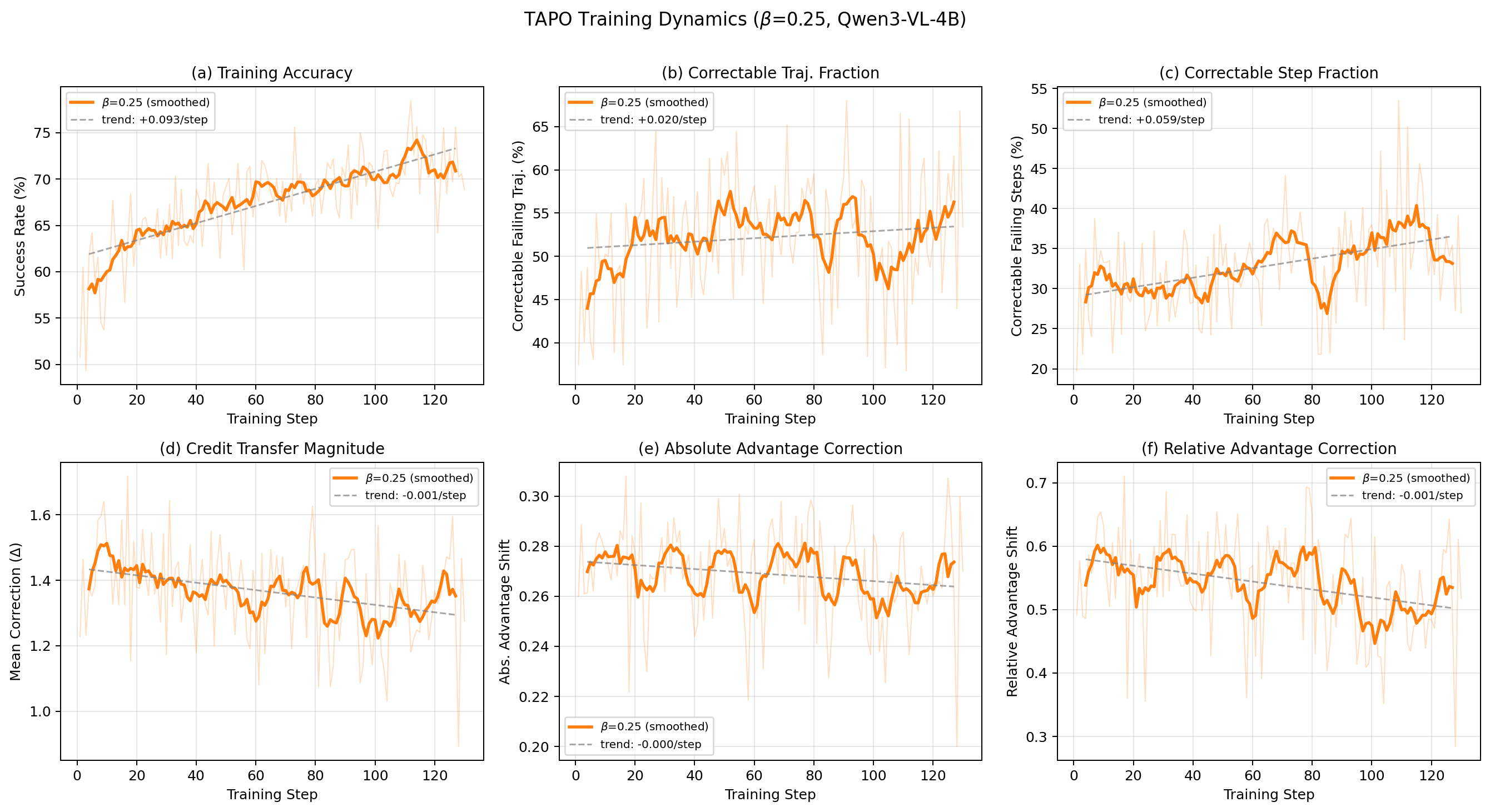}
  \caption{Six-dimensional training dynamics of TAPO ($\beta{=}0.25$, Qwen3-VL-4B).
  \textbf{(a)}~Training accuracy;
  \textbf{(b)}~Correctable trajectory fraction;
  \textbf{(c)}~Correctable step fraction;
  \textbf{(d)}~Mean credit transfer magnitude~$\Delta$;
  \textbf{(e)}~Absolute advantage correction;
  \textbf{(f)}~Relative advantage correction.
  Dashed lines indicate linear trend fits.}
  \label{fig:appendix_tapo_dynamics}
\end{figure*}

Figure~\ref{fig:appendix_tapo_dynamics} presents the complete training dynamics of TAPO at $\beta{=}0.25$.
Training accuracy (panel~a) exhibits a sustained upward trend; meanwhile, the correctable trajectory fraction (panel~b) and correctable step fraction (panel~c) also grow slowly, consistent with the analysis in Section~\ref{sec:appendix_misassign_persist} and further confirming that the demand for compensation does not diminish during training.

The mean credit transfer magnitude $\Delta$ (panel~d) remains broadly stable throughout training without systematic decay, indicating that the advantage correction applied at each compensation step is consistently sized.
Both the absolute advantage correction (panel~e) and the relative advantage correction (panel~f) exhibit minimal drift, demonstrating that TAPO's compensation strength adapts to maintain a stable level throughout training---neither growing excessively as training proceeds nor degrading as the success rate rises.

\subsection{Per-Tool-Type Credit Transfer Analysis}
\label{sec:appendix_per_tool}

\begin{figure*}[t]
  \centering
  \includegraphics[width=\textwidth]{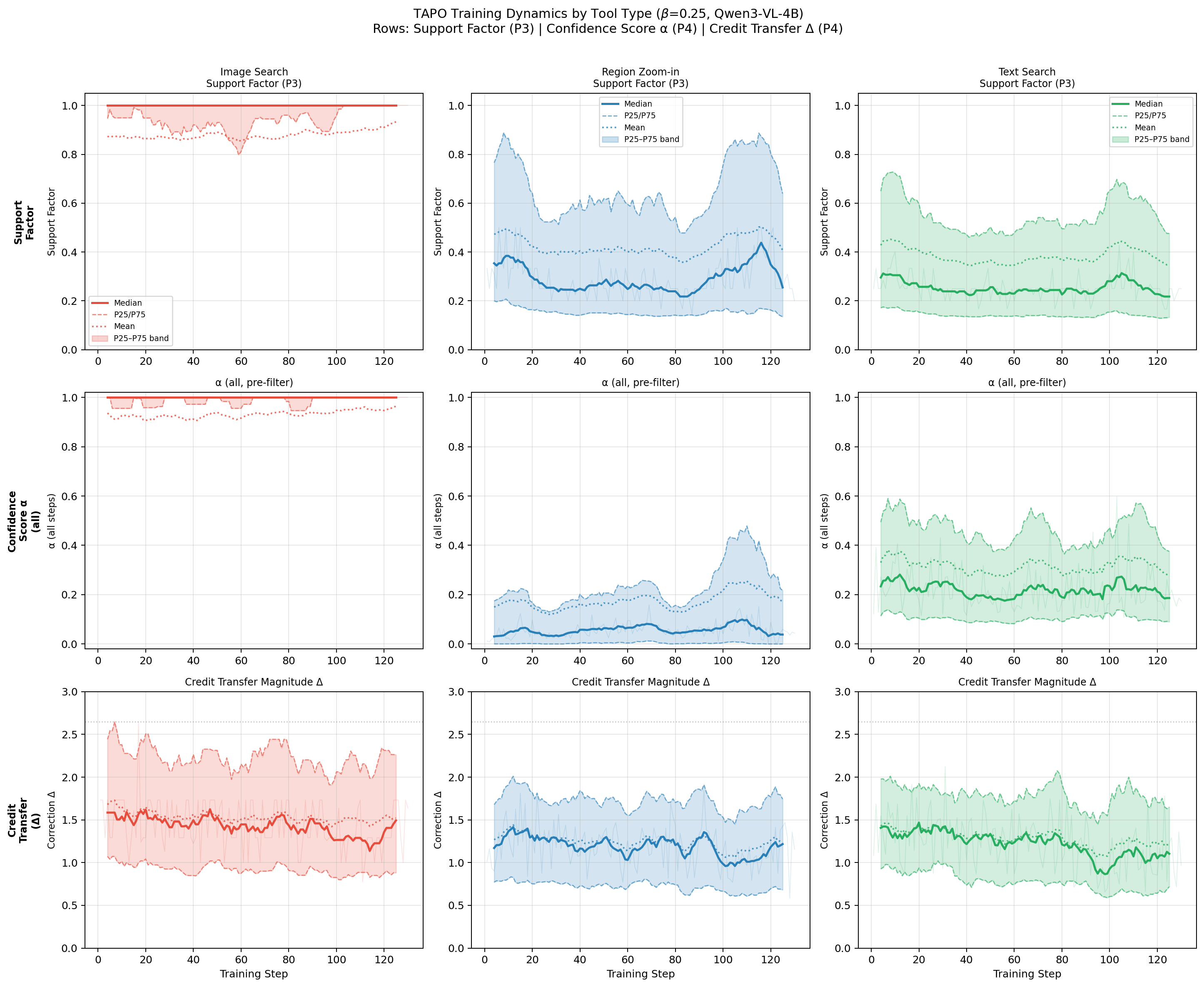}
  \caption{Per-tool-type training dynamics of TAPO ($\beta{=}0.25$, Qwen3-VL-4B) across 130 training steps.
  Columns correspond to three tool types (Image Search / Region Zoom-in / Text Search).
  Rows show the support factor (top), confidence score~$\alpha$ (middle), and credit transfer magnitude~$\Delta$ (bottom), each plotted as median, P25/P75 band, and mean.}
  \label{fig:appendix_per_tool}
\end{figure*}

Different tool types exhibit a pronounced and stable stratification pattern under TAPO correction, which can be understood systematically along three dimensions.

\paragraph{Support factor stratification.}
As shown in the top row of Figure~\ref{fig:appendix_per_tool}, image search achieves a median support factor close to~1.0, with the P25--P75 band highly concentrated near the upper bound.
This property stems from the consistency of image search call parameters: for a given question, all image search steps share identical inputs, so a reference group covering all successful trajectories can almost always be found within the same training batch, maximally satisfying the support condition.
By contrast, region zoom-in and text search exhibit much wider P25--P75 bands and lower mean values, reflecting higher diversity in call parameters across samples and correspondingly limited reference group coverage.

\paragraph{Stable stratification of confidence score~$\alpha$.}
The middle row shows that image search achieves a median $\alpha \approx 0.94$, concentrated in the high-value range, while region zoom-in ($\alpha \approx 0.05$--$0.15$) and text search ($\alpha \approx 0.15$--$0.35$) are substantially lower with wider spread.
This three-tier stratification remains stable across all 130 training steps, demonstrating that TAPO's confidence estimation mechanism reliably discriminates between tool types and does not drift with training state.

\paragraph{Consistency between $\Delta$ and confidence stratification.}
The bottom row reveals that image search has the highest median $\Delta \approx 1.5$, while region zoom-in and text search are somewhat lower (approximately 1.0--1.2), with all three remaining stable throughout training.
The inter-tool ranking of $\Delta$ is entirely consistent with the stratification of $\alpha$, indicating that TAPO's correction magnitude adaptively reflects signal quality rather than applying a uniform fixed correction to all tool steps: stronger corrections are applied to high-confidence image search steps, while more conservative compensation is maintained for lower-confidence text search steps with lower query similarity.

The consistency across all three metrics jointly validates the effectiveness of TAPO's confidence-gated design: through the confidence gating mechanism, TAPO achieves differentiated advantage compensation across tool types, actively correcting high-quality signals while avoiding over-intervention on less reliable ones.

\section{Case Study: Reasoning Patterns of TAPO-Trained Model}
\label{sec:appendix_cases}

To qualitatively illustrate the search strategies learned by TAPO, we present four representative cases from the evaluation set, covering different reasoning patterns exhibited by the trained model.
Each case shows the input question, the ground-truth answer, the sequence of tool calls invoked, and the model's reasoning trajectory.

\begin{figure*}[t]
  \centering
  \resizebox{0.95\linewidth}{!}{\includegraphics{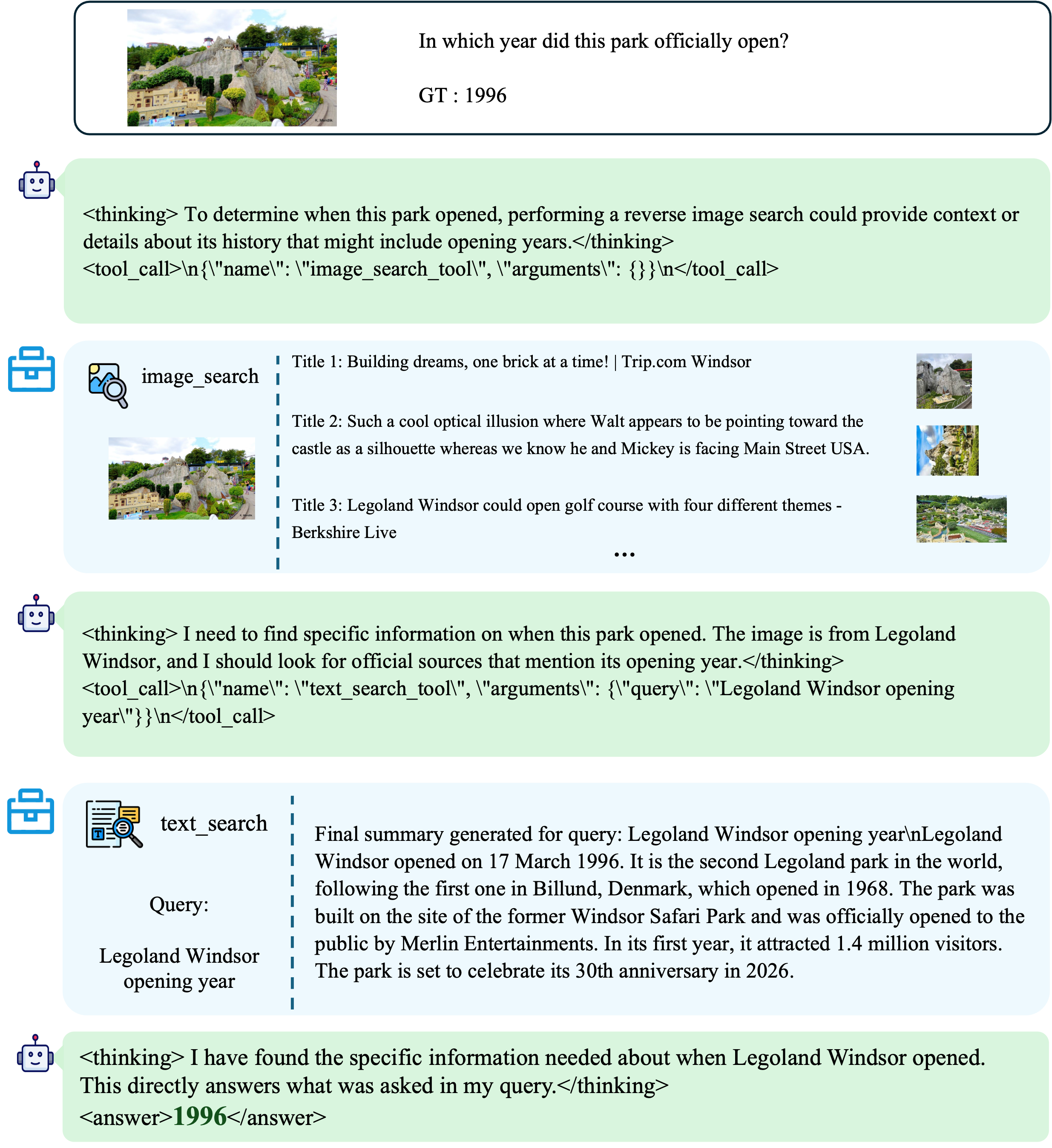}}
  \caption{\small
    \textbf{Case 1: Entity Recognition + Knowledge Retrieval.}
    The model identifies \emph{Legoland Windsor} via image search and retrieves its opening year with a single text query, correctly answering \emph{1996}.
  }
  \label{fig:case1}
\end{figure*}

\paragraph{Case 1: Entity Recognition + Knowledge Retrieval.}
The question asks: \textit{``In which year did this park officially open?''} (ground truth: \textbf{1996}).
The model performs a reverse image search that returns results including ``Legoland Windsor could open golf course'' and ``LEGOLAND Windsor Resort'', enabling it to identify the park.
It then queries \texttt{text\_search\_tool} with \emph{``Legoland Windsor opening year''}, obtaining the precise opening date of 17 March 1996.
This two-step pattern---visual entity identification followed by a single factual lookup---is the most common successful strategy on knowledge-intensive benchmarks such as InfoSeek.

\begin{figure*}[p]
  \centering
  \resizebox{0.90\linewidth}{!}{\includegraphics{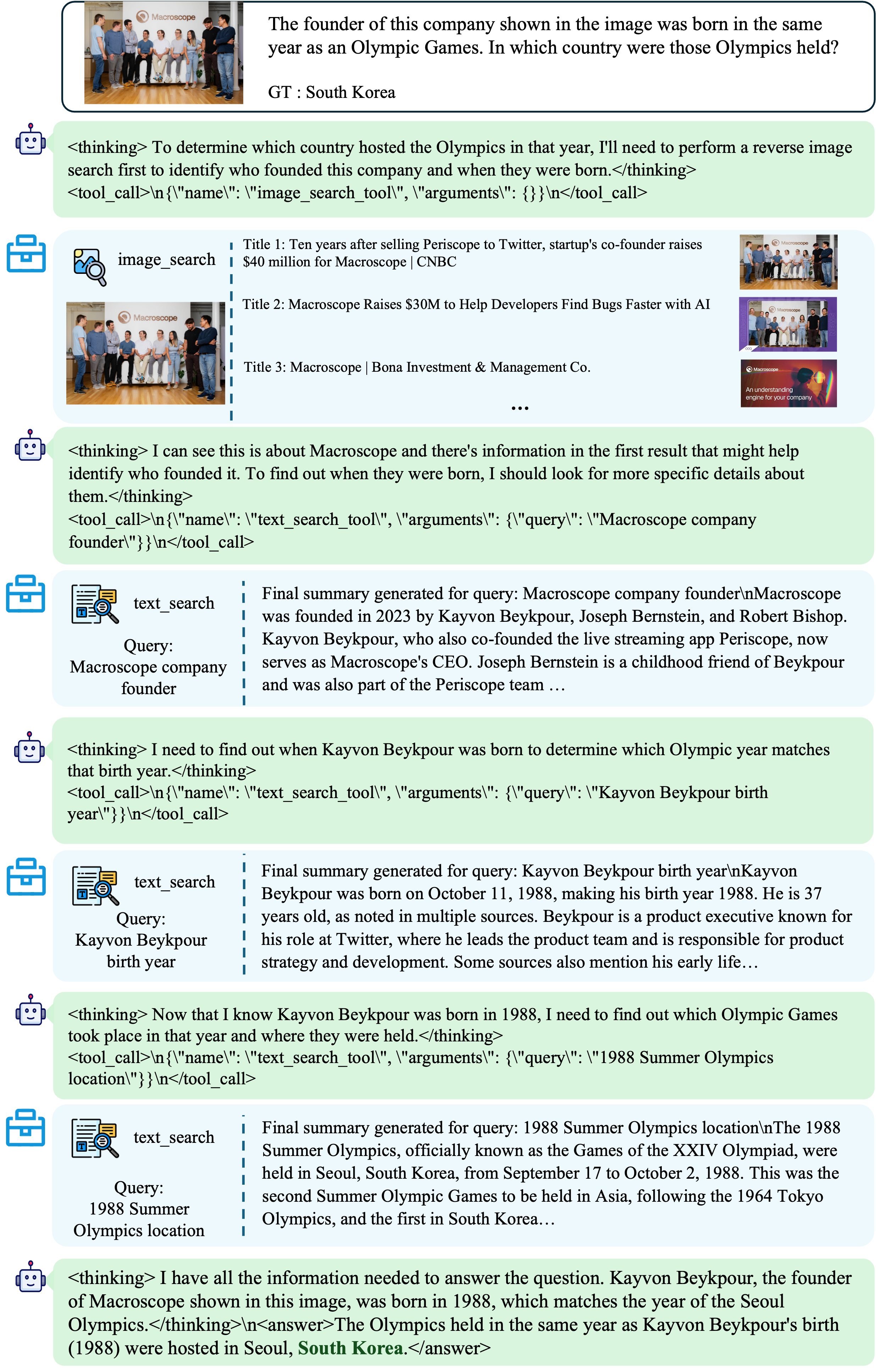}}
  \caption{\small
    \textbf{Case 2: Multi-Hop Iterative Query Refinement.}
    Starting from a group photo, the model chains four tool calls across three reasoning hops to arrive at \emph{South Korea}.
  }
  \label{fig:case2}
\end{figure*}

\paragraph{Case 2: Multi-Hop Iterative Query Refinement.}
The question asks: \textit{``The founder of this company shown in the image was born in the same year as an Olympic Games. In which country were those Olympics held?''} (ground truth: \textbf{South Korea}).
The model employs a four-tool chain spanning three reasoning hops:
(1) \texttt{image\_search\_tool} returns CNBC headlines about \emph{Macroscope}, identifying the company;
(2) \texttt{text\_search\_tool}[\emph{``Macroscope company founder''}] confirms the CEO as \emph{Kayvon Beykpour};
(3) \texttt{text\_search\_tool}[\emph{``Kayvon Beykpour birth year''}] retrieves birth year \emph{1988};
(4) \texttt{text\_search\_tool}[\emph{``1988 Summer Olympics location''}] returns \emph{Seoul, South Korea}.
Each intermediate result directly seeds the next query, bridging visual entity recognition, biographical lookup, and historical knowledge into a coherent reasoning chain.

\begin{figure*}[p]
  \centering
  \resizebox{0.90\linewidth}{!}{\includegraphics{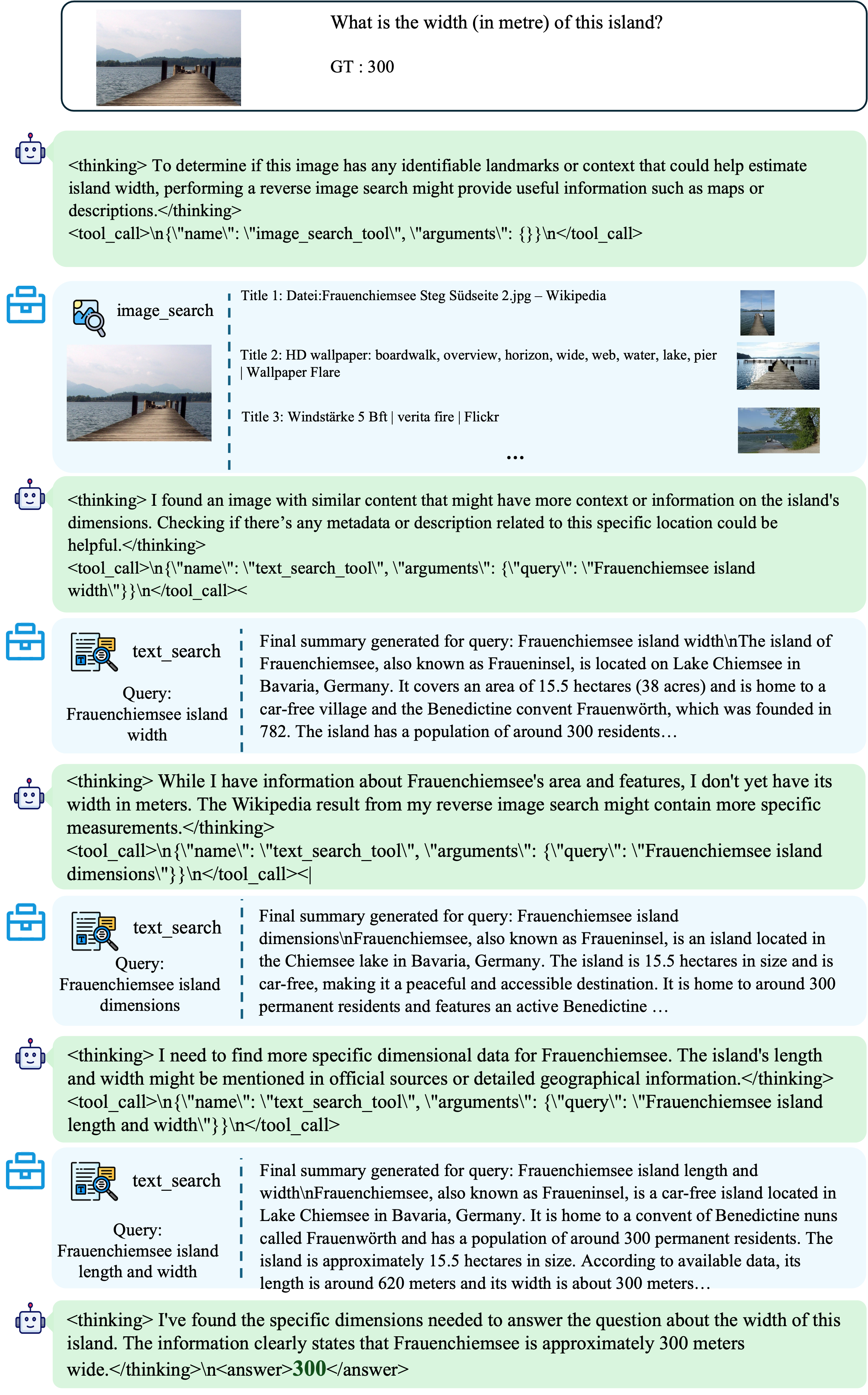}}
  \caption{\small
    \textbf{Case 3: Persistent Query Refinement for Numerical Retrieval.}
    The model refines its text query three times after receiving non-specific results, ultimately extracting the exact island width of \emph{300\,m}.
  }
  \label{fig:case3}
\end{figure*}

\paragraph{Case 3: Persistent Query Refinement for Precise Numerical Retrieval.}
The question asks: \textit{``What is the width (in metre) of this island?''} (ground truth: \textbf{300}).
The model invokes \texttt{image\_search\_tool} and identifies the island as \emph{Frauenchiemsee} from the Wikipedia result.
Three successive text searches are then issued with increasing specificity:
(1) \emph{``Frauenchiemsee island width''} returns area (15.5\,ha) and population, but no explicit width;
(2) \emph{``Frauenchiemsee island dimensions''} similarly omits the metric width;
(3) \emph{``Frauenchiemsee island length and width''} finally retrieves \emph{``approximately 620\,m long and 300\,m wide''}.
This adaptive reformulation---persisting through partial failures until the precise numerical fact is extracted---demonstrates the model's ability to iteratively refine its information-seeking strategy rather than giving up prematurely.

\begin{figure*}[t]
  \centering
  \resizebox{0.95\linewidth}{!}{\includegraphics{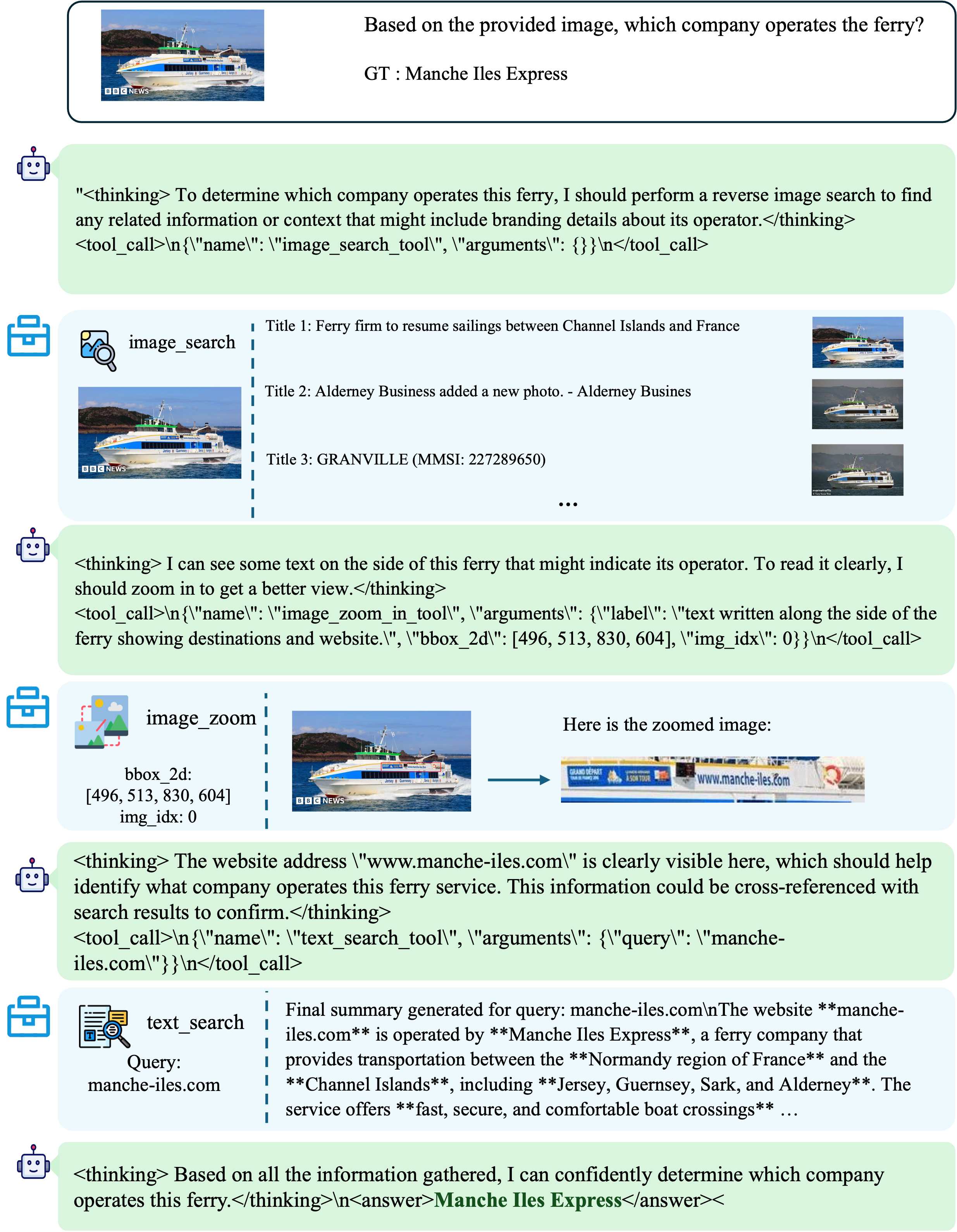}}
  \caption{\small
    \textbf{Case 4: Visual Detail Extraction via Zoom + Text Confirmation.}
    Unable to identify the operator from a global image search, the model zooms into the hull text, reads the URL, and confirms \emph{Manche Iles Express} via a follow-up search.
  }
  \label{fig:case4}
\end{figure*}

\paragraph{Case 4: Visual Detail Extraction via Zoom + Text Confirmation.}
The question asks: \textit{``Based on the provided image, which company operates the ferry?''} (ground truth: \textbf{Manche Iles Express}).
The model first calls \texttt{image\_search\_tool}, which returns ferry-related headlines (e.g.\ ``Ferry firm to resume sailings between Channel Islands and France'') but no direct operator name.
Recognising that the text painted on the hull may be legible at higher resolution, it invokes \texttt{image\_zoom\_in\_tool} on region \texttt{[496, 513, 830, 604]}, successfully reading \emph{``www.manche-iles.com''}.
A final \texttt{text\_search\_tool}[\emph{``manche-iles.com''}] confirms that this URL belongs to \emph{Manche Iles Express}, a ferry service between Normandy and the Channel Islands.
This case exemplifies a \emph{coarse-to-fine} visual strategy: global image search for scene context, zoom for fine-grained textual evidence, and a confirmatory knowledge lookup.

\end{document}